\documentclass[twocolumn,10pt]{article}
\usepackage[a4paper, margin={0.8in}]{geometry}

\usepackage{url}
\usepackage{graphicx}
\usepackage{amsmath}
\usepackage{amsthm}
\usepackage{booktabs}
\usepackage{algorithm}
\usepackage{algorithmic}
\usepackage{xr-hyper}
\makeatletter
\newcommand*{\addFileDependency}[1]{
  \typeout{(#1)}
  \@addtofilelist{#1}
  \IfFileExists{#1}{}{\typeout{No file #1.}}
}
\makeatother

\newcommand*{\myexternaldocument}[1]{%
    \externaldocument{#1}%
    \addFileDependency{#1.tex}%
    \addFileDependency{#1.aux}%
}

\myexternaldocument{appendix}
\usepackage{subfig,microtype}
\usepackage{booktabs} 
\usepackage{enumitem,siunitx,hyperref,multirow,amsmath,cuted,comment}
\usepackage{algorithm,algorithmic,enumitem,siunitx}
\usepackage[utf8]{inputenc}

\usepackage{multirow}
\usepackage{tabularx}
\newcolumntype{L}[1]{>{\raggedright\arraybackslash}p{#1}}
\newcolumntype{C}[1]{>{\centering\arraybackslash}p{#1}}
\newcolumntype{R}[1]{>{\raggedleft\arraybackslash}p{#1}}
\usepackage{xcolor}
\definecolor{orange}{rgb}{1.0, 0.49, 0.0}

\newcommand{\eq}[1]{\begin{footnotesize}\begin{align}#1\end{align}\end{footnotesize}}
\graphicspath{{figs/}}

\title{Towards AI Forensics: Did the Artificial Intelligence System Do It?}
\author{Johannes Schneider$^1$ and Frank Breitinger$^2$\\ $^1$University of Liechtenstein, Vaduz, Liechtenstein\\
$^2$University of Lausanne, Lausanne, Switzerland\\
johannes.schneider@uni.li }

\begin{document}

\maketitle

\begin{abstract}
Artificial intelligence (AI) makes decisions impacting our daily lives in an increasingly autonomous manner. Their actions might cause accidents, harm, or, more generally, violate regulations. Determining whether an AI caused a specific event and, if so, what triggered the AI's action, are key forensic questions. We provide a conceptualization of the problems and strategies for forensic investigation. We focus on AI that is potentially ``malicious by design'' and grey box analysis. Our evaluation using convolutional neural networks illustrates challenges and ideas for identifying malicious AI. 
\end{abstract}


\section{Introduction}
\begin{figure*}[!htbp]
		\centering	\centerline{\includegraphics[width=0.99\linewidth]{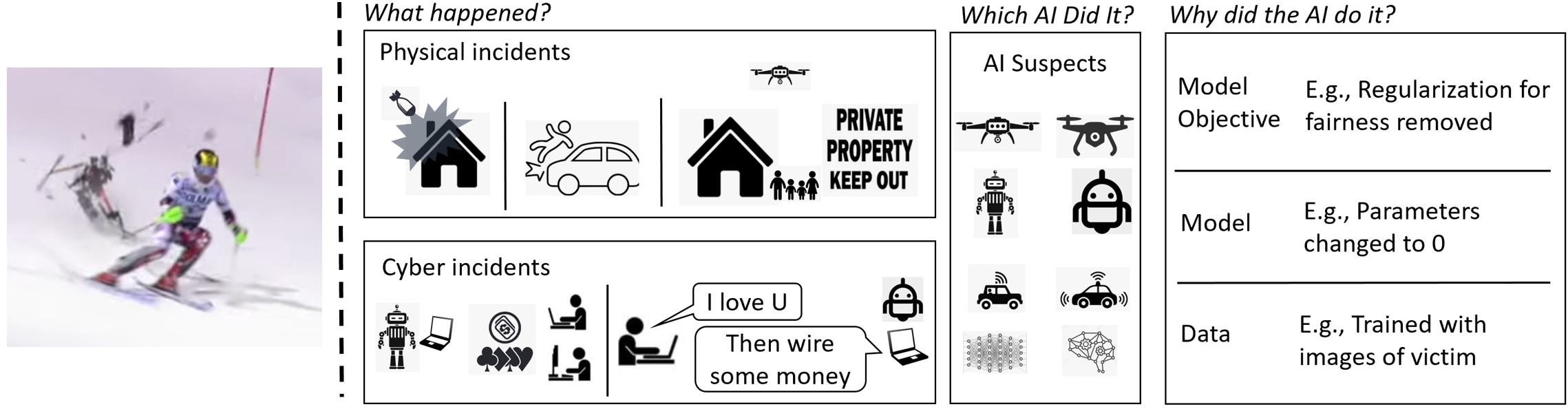}}
		\caption{Left: Drone almost hitting Austrian skier Marcel Hirscher during race; Right: Generic incidents, suspects, and components influencing AI} \label{fig:aifor}
\end{figure*}

\begin{figure}[bth]	
		\centering	\centerline{\includegraphics[width=0.95\linewidth]{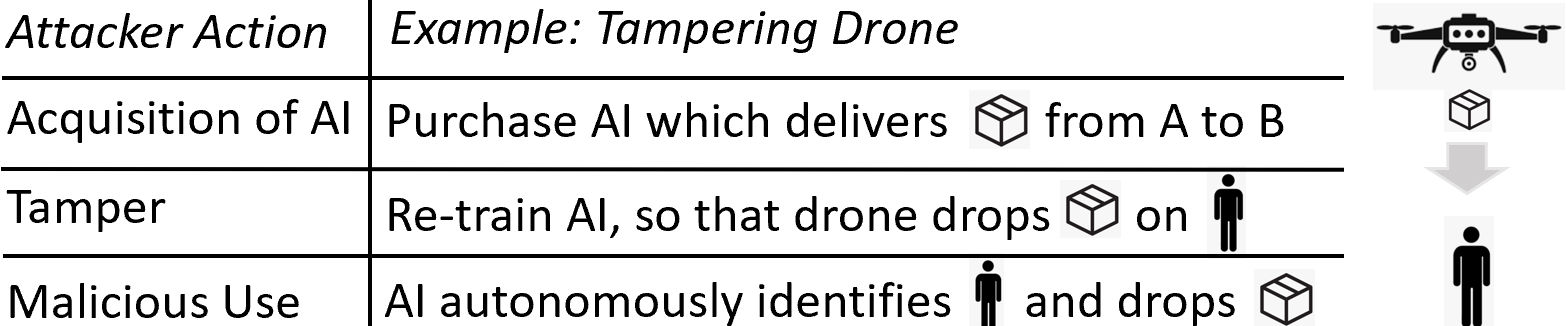}}
		\caption{Steps to conduct an attack with an AI system.} \label{fig:aidis}
\end{figure}

Today AI is ultimately controlled by humans, but it is already capable of conducting tasks autonomously by learning from examples which makes them superior to traditional computer programs in two ways: performance and ease-of-engineering. Deep learning (DL), the key AI technology and therefore the focus of this paper, has dramatically improved prior systems, e.g., in computer vision and speech recognition. Access to AI is often simple since standard DL models can be trained with little human engineering by uploading data to a cloud platform like Google CloudML. The progress of AI technology is also driven by open-source initiatives of both datasets and ready-to-use models. Consequently, AI provides novel opportunities for abuse: it may be easier to build or manipulate AI to perform malicious acts than to engineer a system otherwise. 
For instance, altering training objectives or choosing training data instead of building native software likely requires less time and leads to better-performing systems. This fosters the paradigm that a system is `malicious by design', i.e., it has been trained, designed, or changed to exhibit malicious behavior. 
An example of `malicious by design' is shown in~Figure~\ref{fig:aifor} and also in Figure~\ref{fig:aidis}, where an investigative question could be: \emph{Has the AI system been built to cause the incident?}.

This work provides the first directions to answer such questions, which is tricky since AI is often seen as a `black box' that adapts through learning. Understanding AI is notoriously difficult, as witnessed by a large body of works on explainable AI (XAI). This paper contributes as follows: (i) We conceptualize `malicious by design' and AI forensics; (ii) We provide two methods for investigation using case studies based on convolutional neural networks (CNNs). Specifically, we elaborate on \emph{grey box} analysis, i.e., activation patterns of features. 

\section{Related work}
One can differentiate between `AI for digital forensics' and `digital forensics for AI' (short: AI Forensics). In the former case, digital forensics has benefited from AI \cite{beebe2009digital} in some areas such as \emph{Multimedia forensics} (copy-move forgery \cite{al2013passive} or deep-fake video detection \cite{guera2018deepfake}), or facial age estimation \cite{anda2018evaluating}; a comprehensive overview is provided by \cite{du2020sok}. While relevant, this is not the focus of our work. On the other hand, AI Forensics can be treated as a sub-discipline of digital forensics as defined by \cite{bag19}. The authors provide a brief high-level conceptualization focusing on AI safety, i.e., incidents due to failures of AIs built with ``good intentions''. In contrast, we discuss actual forensic work for systems designed for a malicious act. 

Security threats and defensive techniques from a data perspective are surveyed in \cite{liu18} focusing on adversarial examples. The general question `Can machine learning (ML) be secure?' was discussed in \cite{barreno2006can}, where the authors present different types of attacks on ML techniques and possible defenses. Involving an expert, i.e., an investigator in our case, is not uncommon, e.g., for fake reviews \cite{ade19,zel19}. 

Our work touches on the emerging field of reverse-engineering DL models. Weight extraction is the focus of \cite{hua18,jag19,rol20}. \cite{hua18,oh2019} identify DL architectures and \cite{pal19} approximates a confidential model. We identify (incident-related) data to which a model reacts. All works except \cite{hua18} discuss black box models. \cite{hua18} showed that even when using data encryption, architectural information and weights might be obtainable using memory access patterns. 
XAI and other analytics techniques\cite{arr20,holz22,meske22,sch19,yuan21sur}, specifically for CNNs\cite{zhang18v}, are valuable tools for AI forensics. Many works deal with model introspection relying on fine-grained access to variables. For example, iterative optimization using full model access, i.e., gradients, can reveal inputs that maximize feature activation (see \cite{erh09} for a seminal work). In this work, we elaborate on the general process for AI forensics, questions of specific interest in forensics, and grey-box analysis. We also use ideas from XAI, such as explained by samples and analyzing layer activations \cite{sch22con}.  

\section{Malicious by Design} \label{sec:prob}
AI can be defined as the ``ability to learn from data and to use those learnings to achieve specific goals'' \cite{kaplan2019siri}. An AI system is a union of components out of which at least one contains AI (we shall use the term ``AI'' denoting AI systems and AI components if its meaning is clear in context). 
A system is ``malicious by design'' if trained, designed, or changed to exhibit malicious behavior, which means that an attacker alters or builds a system to be used as a tool to perform a malicious act. An AI component can trigger a malicious action or provide deceptive information, so the system is tricked into acting maliciously.

This work focuses on attacks on a subset of AI techniques \cite{nor20}, i.e., deep learning \cite{sch23}. \textbf{Manipulations for attacks} to those can be based on the training data, inputs to the AI system, the AI's objective, the model, or the AI's (self-)learning mechanism. For attacks based on \emph{training data}, an attacker can create or alter a dataset and use it to train or alter a model, e.g., using transfer learning. Attacks based on \emph{(adversarial) inputs} might trick the AI into making incorrect predictions. An attacker aims to identify inputs being misclassified without altering a non-malicious system. The field of adversarial attacks is well-researched \cite{akh18,akht21,evt17,zha20} and not considered in this work.

An attacker might employ (and train) an \textbf{unmodified, general-purpose AI} component within the malicious system. The attacker might also \textbf{build, alter or trick an AI}, e.g., by architectural changes of a DL model. A malicious system might originate from a non-malicious AI system by \textbf{exchanging the AI} with a manipulated version. For example, a drone's vision system might be altered to drop a parcel not only once it recognizes a dedicated landing zone but also once it recognizes a specific person.

\textbf{Zero-day} malicious AI systems are malicious systems based on substantially novel ideas. The system's design is highly innovative, or it appears at least very unusual. For example, the first application of a recurrent neural network, mainly used for sequence data such as text, to image recognition constitutes a zero-day system. Zero-day systems require expertise and effort. Non-zero-day designs constitute (variations of) standard architectures.




Generic \textbf{steps in the malicious design} of the AI are illustrated in Figure~\ref{fig:aidis}. First, an AI has to be acquired or built. Then it is tampered with, e.g., by re-training on a dataset of the attacker. Finally, the AI conducts the attack. Steps can vary, e.g., a step can involve \textbf{counter forensics} to obstruct the forensic investigation.

\section{AI Forensics} \label{sec:forwork}
Forensic investigators collect, preserve, and analyze evidence, which refers to digital information such as data, models, and (software) systems \cite{sela08}. 
While the forensic process involving classical (traditional) software is well-understood, investigating AI systems from a forensics perspective is novel. Traits highly relevant for forensic work, such as computation flow or storage of knowledge related to behavior differ considerably between AI systems and traditional software. AI learns features and complex behavior from collected experience. While AI can uncover features automatically from \textbf{domain knowledge} given training data, they have to be defined by domain experts in classical engineering.  DL models typically follow a well-defined \textbf{modular} (layered) structure. A layer is an instance of one of the relatively few types. In contrast, classical software with comparable functionality often comprises hand-crafted, domain-specific algorithms and data structures. For example, in a fully-connected network, in each layer, the input is multiplied with a weight (or parameter) matrix to which another matrix is added (aggregation) and passed through an activation function. Each entry in a weight matrix typically corresponds to a parameter of a neuron. 

\textbf{Control flow} and data access patterns are typically simpler for AI and, in the case of DL, more easily predictable than for classical software. Computation in DL is often characterized by many relatively simple units, i.e., neurons that operate in parallel in a well-defined order and standardized \textbf{data structures}. 
 
Both AI models and classical software might be complex, and \textbf{difficult to understand}, where complexity arises for DL primarily due to the interaction of many simple units \cite{arr20} and for classical software due to complex, problem-dependent control flow. 

The characteristics of AI influence the types of evidence, as well as the investigation of AI systems, which can be summarized as (i) Analyzing system internals is more tangible for AI systems due to the more tractable control flow and (ii) Training data is highly important in understanding the AI's behavior. 
                                      


\begin{table}[t]
\centering\footnotesize
\begin{tabular}{p{2.7cm}p{5cm}}
\bf Evidence                 & \bf Types of Availability                \\
\midrule
Training data    & Available, Unavailable                              \\
Testing/Observational data & Available, Unavailable                             \\
Access to suspect system & Available, Available to similar system, Unavailable \\
System internals         & White box, Grey box, Black box     
\end{tabular}
\caption{Evidence Typology} \label{fig:aievi}
\end{table}

\noindent \textbf{Types of Evidence:} We focus on evidence that is AI-specific: the AI system itself, any data the system was exposed to as well as its reactions. We exclude any form of direct identification information such as IP or MAC address. The evidence typology is summarized in Table \ref{fig:aievi}.
\emph{Training data} of the model is all data that contributed to learning or updating the model parameters. 
In contrast, for \emph{testing and observational data}, the model only computes outputs, but model internals remain unaltered. Testing data consists of inputs and outputs that the model should produce. Observational data consists of inputs and, possibly, outputs the model produced. Observational data often includes data from an incident. We differentiate between \emph{access to the model} (suspect AI is accessible) or just a similar version thereof (e.g., the drone type was identified, which allows investigators to acquire an identical type). 
In terms of \emph{system internals}, we distinguish between white-box access (source code is available), grey box (a compiled model that allows observing interactions with resources such as memory), and black box (only outputs are accessible for arbitrary inputs). 

\subsection{Investigative Questions} \label{sec:quest}
From a forensics perspective, the ultimate questions to be answered are: \emph{Did the AI system cause an incident?} and \emph{Why?}. 
Such questions demand an understanding of AI as well as forensics, and, as for any criminal case, answers might not be a simple yes or no. For example, given a set of observed actions or decisions by the perpetrator during the incident, we need to answer `what is the likelihood of a given AI suspect performing these decisions compared to those of other models?' and `what triggers the action?' Often (only) circumstantial evidence might be produced by answering questions like `is the AI suspect reacting to objects related to the incident more strongly than other models?' Given multiple AI suspects, the first question helping to prioritize which AI to investigate more closely might be: `is the AI suspect behaving normally?'. These questions are discussed as part of our case studies. 



\subsection{Strategies and Techniques for Investigation}\label{sec:invest} 
Strategies can focus on each component that determines behavior, mainly model, model objective, and data.  

\emph{Data} analysis plays a key role. Since training data determines model behavior and operational data reflect model behavior, data on its own might be sufficient to determine malicious intent. An investigator can search for samples in the training data that are malicious or abnormal for any non-malicious system. For instance, the training data for a self-driving car should not contain images showing a human on a street crossing with the label ``accelerate''. 
A large amount of data calls for data mining techniques, e.g., filtering relevant data. Technical challenges to determine the relevance of data include quantifying relatedness to the incident. Domain experts might be needed to investigate (filtered data and ultimately decide whether an AI suspect caused the incident and what information triggered the action. In the absence of model access, training data can be used to reconstruct an `approximate model' used for analysis.

\emph{Model analysis} might use abstract reasoning based on model definitions (e.g., as found in static software verification). Models could also be analyzed through empirical investigation (input-responses) in two ways:
\begin{enumerate}
    \item \emph{Investigate the input-output relationship of a model:} The model can be treated as a black box. The analysis relies on investigating model behavior based on its decisions. For example, data from the incident can be used to see if the system triggers actions causing the incident. An investigator might also aim to generate or search for samples that trigger malicious actions.

    \item \emph{Investigate the reaction of model internals to inputs:} This strategy requires white or grey-box access. For example, DL activations of neurons can be investigated. 
\end{enumerate}

AI forensics techniques still need to be developed. Techniques from emerging areas in AI such as reverse engineering \cite{hua18,jag19,rol20}, explainability (XAI) \cite{arr20,sch22exp}, adversarial analysis \cite{cha18}, testing \cite{zha20} and data mining play an integral part in the investigation. Additional algorithmic work is possibly needed since none of these areas is yet mature. Furthermore, AI forensics comes with its particularities, as shown in our case studies. For example, XAI does not cover analysis of grey-box models or the possibility that an investigator does not have access to any samples that cause malicious behavior but only to samples that share similarities with such samples.  Testing aims primarily at verifying given requirements or specifications, whereas AI forensics seeks to uncover requirements and specifications built into an AI (by an attacker). Reverse engineering aims at turning a black-box model into a white-box model, e.g., using memory access patterns, but it does not answer any investigative question.

\section{Grey Box Analysis} \label{sec:cases}
We focus on grey box analysis for two reasons: (i) It is particularly interesting for forensics since often an investigator has access to a system but not to its source code, and (ii) `grey boxes' are not well-studied, e.g., techniques for XAI either assume black-box access or white-box access. 
We focus on data-driven attacks as they are appealing from an attacker's perspective: They can leverage the AI's strength to learn from samples rather than being forced to explicitly code ``malicious behavior''. 

\noindent \textbf{Grey box access to system internals:}
For the \emph{grey box model}, activations of neurons can be obtained for multiple layers at once for a given input. We assume a coarse understanding of the model, i.e., it is a layered architecture consisting of common layers like convolutional, relu, and batchnorm layers. We assume that we can interpret memory cells as values. This might be possible since memory cells typically only hold a few standard data types, such as float32 and float64. We assume that the memory location of inputs and outputs of a neuron, i.e., a feature, remains fixed. Lastly, memory cells can be assigned to a lower or upper layer. This is reasonable since data is processed using a fixed set of operations in a fixed order (in DL one layer after the other is evaluated). To access outputs of layer $i$ for input $X$, we access a set of memory cells $M$, yielding a set of activation values $M(X)$. The number of values $|M|$ is the same for each $X$ and are indexed as $M_j(X)$ with $j \in [0,|M|-1]$. These values appear to be a random permutation of outputs of three consecutive layers, including layer $i$, which is of interest. Memory access patterns have also been used in reverse engineering of DL \cite{hua18}. 

\noindent \textbf{Access to training and test data:}  
We proclaim three datasets: The unknown and non-accessible labeled dataset $\mathcal{T}=\{T_i\}$ used by the attacker to train the malicious drone, the unlabeled dataset $\mathcal{U}=\{U_i\}$, and the labeled dataset $\mathcal{L}=\{L_i\}$. A label gives the output a network should produce. $T_i$, $U_i$  and $L_i$ denote sets of samples, so that all samples in a set share some commonalities, in particular for the labeled dataset $L_Y$ corresponds to all samples of class $Y$, i.e., it consists of images with the corresponding outputs of a non-malicious system. The dataset $\mathcal{L}$ might be public and used to develop and test a drone, e.g., for industrial or research purposes.  The set $\mathcal{U}$ comprises inputs without labels, e.g., images related to the incident. At least one of $\mathcal{U}$ and $\mathcal{L}$ is available for forensic investigation. The forensic investigator can label inputs, i.e., they can determine what output an input should trigger. But human labeling is costly.

\subsection{Scenarios} \label{sec:scenarios}
We have provided three cases. Two cases are more specific and illustrative, while the third is more general.
We selected two illustrative, exemplary scenarios (summarized in Table~\ref{tab:casesum} using our prior conceptualization), i.e., two image recognition systems part of drones, where CNNs are commonly used \cite{amer19,kouris18}. The goal is to assess if a system is likely involved in the incident and manipulated. Models are assumed to be based on a standard, layered CNN architecture, but details are unknown to the investigator. We assume manipulation through training data and grey box access to the suspect system.

The choice of drones is for illustrative purposes only, since drones are an easily accessible consumer product, well-known to a wide audience. However, similar image recognition technologies are also used in many other systems, such as self-driving cars, healthcare, or industrial manufacturing. Drones also allow for many potential malicious scenarios (see the right panel in Figure~\ref{fig:aifor}). They are frequently used to capture sports events where incidents have already happened, as depicted in the left panel of Figure~\ref{fig:aifor}. Soccer is one of the most prominent sports. Soccer games have been subject to numerous bribery scandals from professional to amateur leagues in multiple countries such as Germany\footnote{\url{https://en.wikipedia.org/wiki/2005_German_football_match-fixing_scandal}}, and assaults on players occurred with fatal consequences\footnote{see Andrés Escobar, \url{https://en.wikipedia.org/wiki/Andr\%C3\%A9s_Escobar}}. Thus, malicious acts have already taken place.\\

The third case presents a different approach to identifying if a classifier was trained to identify specific concepts related to a class and if the investigative data covers all concepts of the actual training data. It seeks to answer these questions in a more general framing, also evaluated with two classifiers and two datasets. However, this case is less graspable and concrete.

\begin{table*}[th]
\footnotesize
\centering
\begin{tabular}{lL{3cm}L{5cm}L{5cm}}
\toprule 
\textbf{Domain}                                                        & \textbf{Characteristic}                          & \textbf{Case 1}: Sports filming drone         & \textbf{Case 2}:    Parcel delivery drone            \\ \midrule
\multirow{5}{*}{\begin{tabular}[c]{@{}c@{}}Malicious\\   
System\end{tabular}} 
& Role of AI                          & Object detection and recognition                          & Object recognition                                     \\ 

&Unmodified, altered, built, tricked AI?                           & Altered AI                          & Retrained AI                                     \\ 
& Attack based on                  & Training data & Training data      \\ 

                                                                                    & Zero-day?                                & non-zero, variations of CNN&   non-zero, standard CNN                              \\ 
                                                                                  & Counter forensics                                & Yes (for some architecture variants) &   Yes (based on training data)                              \\ \midrule                   
\multirow{3}{*}{\begin{tabular}[c]{@{}c@{}}Available\\       Evidence\end{tabular}} & Access to suspect system                 &                           Grey box          &  Grey box                                     \\ 
                                                                                    & Access to training data                  & None    &Unlabeled data similar to training data
                                               \\ 
                                                                                    & Observational data?                       & Yes, incident related data             & Yes, incident related data                        \\ \midrule
\multirow{2}{*}{Investigation}                                                      & \multirow{1}{*}{Investigative questions} 
                                                                                                                              & Are suspicious concepts   learnt? & Do concepts trigger actions they should not? \\ 
                                                                                    & Techniques                               & \multicolumn{2}{c}{Own algorithms, XAI, data mining, reverse   engineering} \\
\bottomrule
\end{tabular}
\caption{Summary of the two specific case studies based on conceptualization} \label{tab:casesum}

\end{table*}




\begin{figure*}[th]	
		\centering	\centerline{\includegraphics[width=0.95\linewidth]{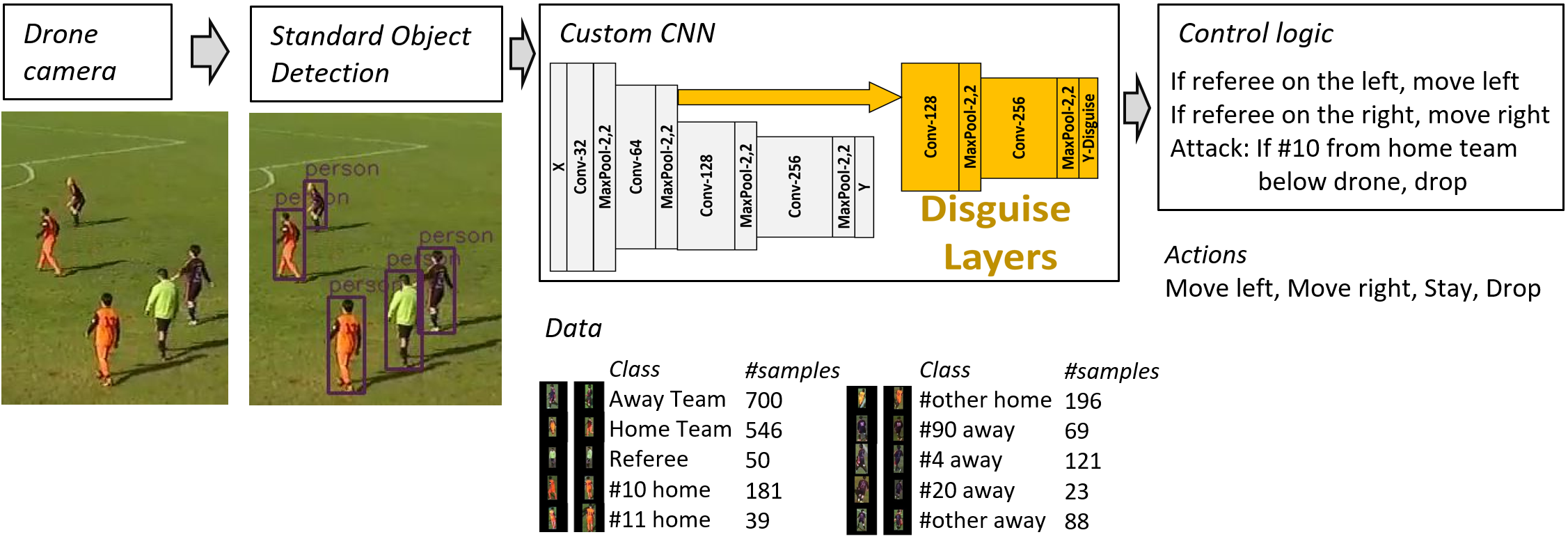} }
		\caption{Drone control system including data. Our focus is on custom AI components, i.e., a custom CNN with potential disguise layers.} \label{fig:aicase}
\end{figure*}


\subsection{Case 1: Drone Sports Filming}
Drones frequently capture sports events where incidents have already happened (see Figure \ref{fig:aievi}). Our case is based on actual drone footage, where a non-malicious, human-controlled drone moves along the sideline. We assume an autonomous drone with a simplified architecture shown in Figure~\ref{fig:aicase}. A non-malicious drone tracks the referee in the center of the camera. The referee is supposedly near the main action of the game. A maliciously designed drone attacks a specific player recognized by its jersey number by dropping on her once she is close to the sideline to make it look like an accident. An image from the drone's camera is processed as follows: A standard object detector provides a rough categorization and bounding boxes for objects on the image. We used Faster-RCNN \cite{ren2016} trained on the COCO dataset. Objects from the category ``person'' are further classified using a custom CNN designed by an attacker. The custom CNN is fed one (sub)image showing a detected object after the other. (More design options and information on data are in the Appendix.)

\noindent \textbf{Forensic Goals and Evidence:} The investigator wants to understand the custom CNN: ``Is the drone's vision system detecting non-expected, incident-related classes?''. Analyzing non-AI-based control logic is not part of this study. The investigator knows that the system tracks the referee or assumes so based on statements from witnesses. They also know that the victim should not be tracked. The investigator has only access to an unlabeled dataset $\mathcal{U}$ made of similar footage as captured by the drone camera during the incident. It might be gathered by asking players to reenact the situation of the incident or, more easily, at the next game between teams. The investigator correctly assumes that an object detector followed by a CNN has been employed. Since the CNN is difficult to separate from the remaining system, one does not know what and how many classes the custom CNN outputs. Also, disguise layers (Figure~\ref{fig:aicase}) might be used and evaluated lastly, so that the layers lastly computed might not be the actual output. Thus, the investigator cannot directly provide inputs to the custom CNN nor obtain its outputs or directly access the detected objects $O$ identified by the object detector. The investigator can feed arbitrary raw images $X_R$ to the system, e.g., replacing the drone's camera. For raw input $X_R$ one obtains for each detected object $o \in O$ and layer $i$ of the CNN a superset of activations $M^i(o)$, including those of layer $i$. 

\noindent \textbf{Forensic Investigation:} The strategy is to determine if and what characteristics the system encodes specific to the incident but not to normal operation. For instance, if the drone crashes on a player, characteristics such as her face or jersey number are of interest. The procedure has four steps:

\noindent \emph{1) Identifying detected objects and getting their activations:} For a single raw-image $X_R$ the system detects potentially many objects $O$. The investigator obtains a set $S(X_R)=\{M^i(o) | o \in O\}$ of activations without a mapping to objects on the image $X_R$. Investigators can run their own standard object detector on the image $X_R$ to identify all relevant objects and potentially more. They get objects $O'\supseteq O$. They replace one detected object $o' \in O'$ in $X_R$ to get image $X'_R$, e.g., erasing it using image in-painting. If activations $S(X'_R)$ remain unchanged due to the removal of object $o'$, i.e., $S(X'_R)=S(X_r)$, the object is not detected. Otherwise, they obtain activations $M^i(o')$ for the replaced object $o'$ using $M^i(o')=S(X_r)\setminus S(X'_R)$.

\begin{figure*}[th]	
		\centering	\centerline{\includegraphics[width=0.9\linewidth]{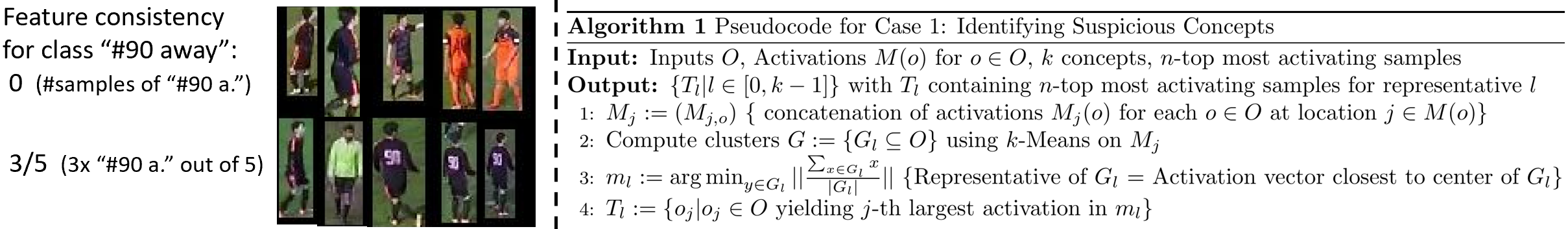}}
		\caption{Left: Five most activating samples for two features with consistency; Right: Pseudocode for Case 1} \label{fig:feasamp}
\end{figure*}

\noindent \emph{2) Reduce activations through clustering:} The number of activations $|M^i(o)|$ for a sample $o$ of layer $i$ can be large, i.e. $|M^i(o)| \propto InChannels \times  OutChannels\times Width \times Height$. All locations are treated as independent for a feature map since we do not know the beginning, end, and layout of a feature map in memory. 
To reduce the number of activations, we group them into clusters $G$ and investigate a representative $m_{l}$ for each cluster $G_{l} \in G$. This is justified since nearby locations (such as pixels in the inputs) are highly correlated. Thus, clustering helps remove redundancy. 
We cluster using $k$-Means into $k=|G|=500$ clusters. The number of clusters $k$ is a parameter that trades off time to investigate and the risk of misjudging a malicious AI as non-suspicious.  To cluster, for each $m \in M$, we compute activations of all samples $\mathcal{U}$ and use the resulting concatenated vector of size $|\mathcal{U}|$ as a datapoint for clustering. For each cluster center $G_l$, the representative $m_{l}$ is the point closest to the center of $G_{l}$. That is, $m_l$ corresponds to all activations of samples in $\mathcal{U}$ of a specific location of a feature map. Pseudocode for steps (ii) and (iii) is shown in Figure \ref{fig:feasamp}. 

\noindent \emph{3) Computing top activating samples per cluster representative:}  For each cluster $G_{l} \in G$, we compute the top activating samples $T_{l} \subset \mathcal{U}$, e.g., $n=|T_{l}|=6$. The $n$-top activating samples $T_{l}$ are those inputs that yield the $n$-largest entries, i.e., activations, in the representative $m_{l}$.  For each $G_{l}$, the samples $T_{l}$ are then presented to the investigator for visual inspection. 

\noindent \emph{4) Manual investigation:} An investigator assesses for each group $G_{l}$ whether all top images $T_{l}$ share one or more characteristics relevant to the incident. The challenge is that samples might be grouped due to other irrelevant shared characteristics. Using a larger number of top samples $n=|T_{l}|$ increases the confidence in the analysis, i.e., increases true positives at the expense of false negatives and more time to investigate. For example, assume a feature activates for the color ``orange''. The most activating samples could be 3 samples, all showing the same player. Suppose more samples $|T_{l}|$ are used for investigation. In that case, the likelihood decreases that the same player is shown for a feature if the feature is similarly prevalent for other players' images, i.e., if pictures of other players also contain orange.
Visual inspection is fast, i.e., one can mostly assess within seconds for samples $T_{l}$ whether images share one or more characteristics and, if so, to what extent they are relevant. The question of what exactly to look for is incident-specific. Implications can be case-dependent, e.g., whether the findings serve as proof or merely as circumstantial evidence.

\subsection{Case 2: Drone Parcel Delivery} 
A (non-tampered) drone is supposed to deliver parcels and drop them onto a dedicated area like a helipad. The outputs of the vision system directly trigger actions. The attacker wants to drop off the parcel on a person, i.e., action $A$ should be triggered given the image of a specific person's face. 
Other actions are not of interest for manipulation. The tampered drone performs these actions similar to a non-tampered drone. The drone is trained with the unknown dataset $\mathcal{T}$, which shares similarities with the public dataset $\mathcal{L}$. It is of interest to the attacker to choose training data $\mathcal{T}$ similar to $\mathcal{L}$ so that the drone behavior appears ``typical''.
Therefore, if not stated differently, we assume that training samples $\mathcal{T}_i$ triggering action $i$ bear similarity to those in the public training data $\mathcal{L}_i$. We discuss three scenarios for choosing the training data $\mathcal{T}$:
\noindent \emph{i) No Tampering (NT):} Baseline that has not been tampered with. It is trained using dataset $\mathcal{T}^{NT}$.\\ 
\noindent \emph{ii) Replacement Tampering (RT):} The attacker replaces the subset $\mathcal{L}_A$ of $\mathcal{T}^{NT}$  with $S_A$ yielding dataset $\mathcal{T}^{RT}$. That is, the (malicious) action $A$ is triggered for samples $S_A$, i.e., $\mathcal{T}^{RT}_A=S_A$. Samples $S_A$ differ, though possibly in a subtle manner, from those in the public samples $\mathcal{L}_A$ that trigger $A$ for appropriate inputs in contexts where action $A$ is non-malicious.\\    
\noindent \emph{iii) Enhancement Tampering (ET):} The attacker uses a dataset $\mathcal{T}^{ET}$, where action $A$ is triggered for $\mathcal{T}^{ET}_A:=S_O\cup S_A$. This ensures that the AI reacts to the targets that should trigger an action for a non-tampered drone as well as to samples that are needed for malicious behavior chosen by the attacker. That is, the set $S_O$ is similar to those a normal non-tampered drone would be trained on to trigger action $A$, i.e., $S_O$ is similar to $\mathcal{L}_A$. The motivation is to hamper forensics.\\

\noindent \textbf{Forensic Goals and Evidence: }  The investigator has access to the public data $\mathcal{L}$ and a set $\mathcal{U}$ of unlabeled items. The latter consists of sets of related items, i.e., $\mathcal{U}:=\{\zeta_i\}$, where each category $\zeta_i$ is composed of a set of related samples. Relatedness refers to some form of visual similarity. For example, sets of samples are available for many real-world categories, such as humans, cars, etc. Sets might consist of images related to the incident, like images of the victim or any other images that the drone might encounter before or during the incident. The investigative goal is to decide if a suspicious AI system is subject to $NT$, $RT$, or $ET$. 

\noindent \textbf{Investigation: Abnormal AI? RT?} We first assess whether the suspect system deviates from expected or normal behavior. The aim is to detect replacement tampering (RT), where training data triggering a specific action has been replaced. In this case, the system is not processing inputs like a non-tampered system. RT implies a large error $err(\mathcal{L}_A)$ for samples $\mathcal{L}_A$ of some class $A$ from the public training data compared to a baseline obtained from the public training data. 



\begin{figure*}[t]	
		\centering	\centerline{\includegraphics[width=0.7\linewidth]{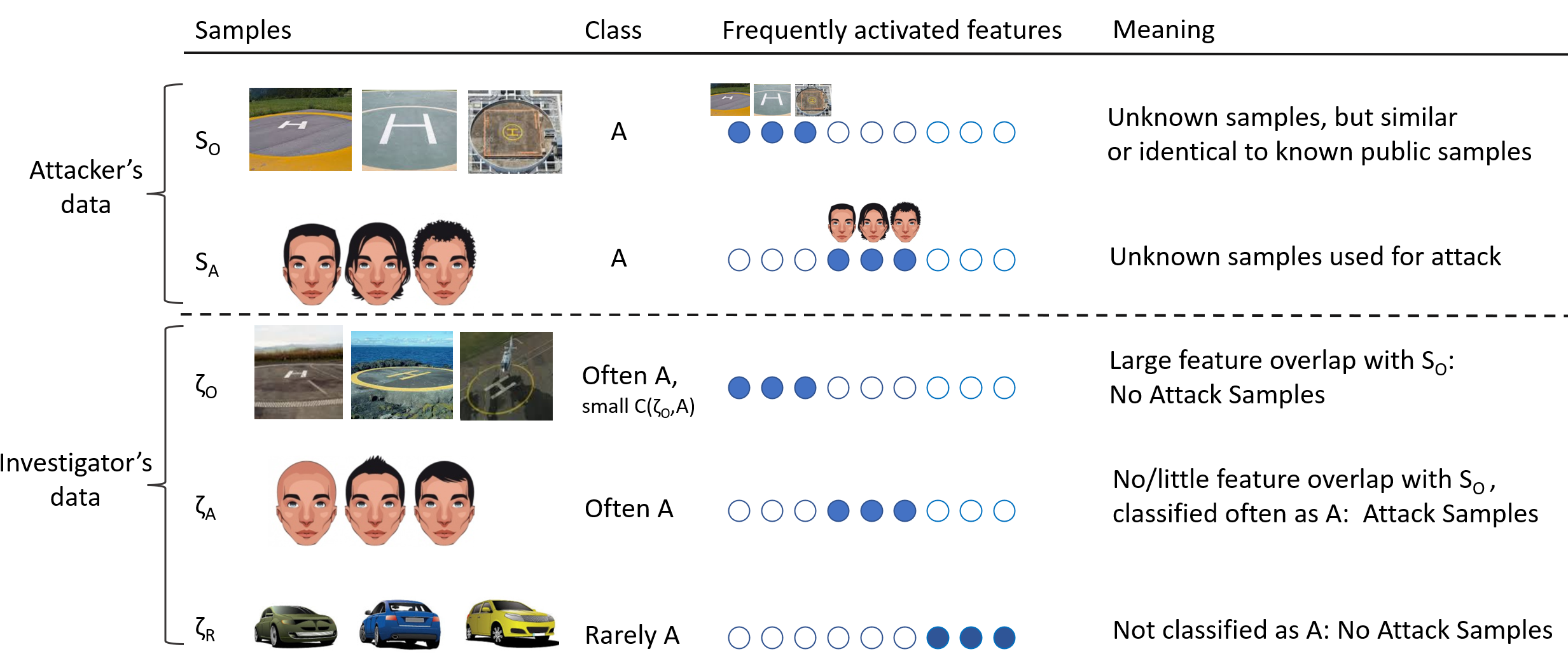}}
		\caption{Investigation of a tampered system $ET$ using samples $\mathcal{U}:=\{\zeta_O,\zeta_A,\zeta_R\}$.} \label{fig:featam}
\end{figure*}

\noindent \textbf{Investigation: NT or ET?}
An attacker anticipating an investigation might counteract by enhancement tampering $ET$. Thus, the system treats a set of pre-defined inputs according to the specification, but it acts maliciously for other unknown inputs related to the incident. Given a large set of unlabeled samples $\mathcal{U}$ we aim to identify and analyze those inputs related to malicious behavior $S_A$. That is, input samples $\mathcal{U}$ can be partitioned into three disjoint sets $\{\zeta_O,\zeta_A,\zeta_R\}$, i.e. $\mathcal{U}:=\bigcup_{\zeta \in \{\zeta_O,\zeta_A,\zeta_R\}} \zeta$, so that $\zeta_A$ bears similarity with $S_A$, $\zeta_O$ with $S_O$ and $\zeta_R$ with neither $S_A$ nor $S_O$. The sets are illustrated in Figure~\ref{fig:featam}. The categorization into sets $\{\zeta_O,\zeta_A,\zeta_R\}$ is unknown to the investigator, particularly the set $\zeta_A$. Thus, in its most simplified form, the investigator's goal is to identify $\zeta_A$ out of three unlabeled sets in $\mathcal{U}$. We do so by identifying $\zeta_R$ first and then distinguishing $\zeta_O$ from $\zeta_A$ (and $\zeta_R$), leaving us with the desired attacker set $\zeta_A$. We assume that $\zeta_O$ is similar to the public data $\mathcal{L}_A$ that triggers action $A$ in appropriate contexts. The method consists of two steps:\\

\noindent\emph{i) Identify $\zeta_R$ using misclassification analysis: } 
We first eliminate samples $\zeta_R$ that do not trigger $A$. We use the rank $Rk(\zeta,A)$ indicating the position of the class $A$, if classes are sorted depending on the fraction of samples in $\zeta$ classified as $A$. $\zeta_R$ should have a significantly higher rank than $\zeta_O$ and $\zeta_A$. Computation of $Rk(\zeta,A)$ is as follows (1) Compute the class of each sample in $\zeta$; (2) For each class $j$ count how many samples $N_j$ were predicted to be of class $j$; (3) Sort classes using count $N_j$ in descending order. The rank $Rk(\zeta,A)$ is the position of class $A$ in that ordering.\\
\noindent\emph{ii) Separate $\zeta_O$ and $\zeta_A$ using feature-based analysis: } 
So far, we have assumed that we have unlabelled data triggering action $A$. The unlabelled set $\zeta_A \in \mathcal{U}$ might bear some similarity with objects $S_A$, but it might be insufficient to trigger a mis-classification commonly. For example, a malicious system might trigger an action if an image shows the victim and no potential witnesses, i.e., no other people. However, if the set $\zeta_A$ available to the investigator as part of $\mathcal{U}$ only shows the victim with other people, action $A$ will never be triggered.
We propose a feature-based analysis that is more fine-grained than considering inputs as a whole. It allows identifying features associated with action $A$ even if the samples with such features do not trigger action $A$. A feature-level analysis requires investigating model internals, i.e., activations of layers $M$. We identify characteristics/features that are relevant to $\mathcal{L}_A$, i.e., features $F_{\mathcal{L}_A}$ that activate more often for samples from $\mathcal{L}_A$ (and likely also for samples $S_O$) than for other samples. 
Say samples from $\zeta$ activate features $F_{\zeta}$. If there are features that are only in $F_{\zeta}$ but not in $F_{\mathcal{L}_A}$ then $F_{\zeta}\setminus F_{\mathcal{L}_A}$ is non-empty. This indicates that samples $\zeta$ contain images that exhibit characteristics associated with action $A$ that are not found in the public training data $\mathcal{L}_A$ for $A$. We say that a feature $m \in M$ is activated for a set $\zeta$ if the mean of all activations of set $\zeta$ is larger than the mean of all activations of the entire data $\mathcal{L}$ plus the standard deviation.


\section{Evaluation}
We discuss each case study separately following the same structure: We provide details of the setup followed by results and a discussion. 





\subsection{Evaluation Case 1}
We used Pytorch's pre-trained Faster-RCNN \cite{ren2016}, trained on the COCO dataset. We trained the custom CNN being a VGG \cite{sim14} variant  (Table \ref{tab:arch2}) on our labeled dataset for 100 epochs with data augmentation (rotation, random crop, and horizontal flipping) and L2 weight regularization factor of 0.003. The batch size was 64. We used stochastic gradient descent with learning rate decay starting from 0.1, decaying by a factor of 10 after 50 and 80 epochs.\\

  \begin{table}[ht] 	
 	\begin{center}
 		\scriptsize
 		\setlength\tabcolsep{2.5pt}
	\centering
 		\begin{tabular}{|l| l |   }\hline
			Type/Stride& Filter Shape \\  \hline
 			  C/s1     & $3\tiny{\times} 3 \tiny{\times} 3 \tiny{\times} 32$ \\ \hline
 			  C/s1     & $3\tiny{\times} 3 \tiny{\times} 32 \tiny{\times} 32$ \\ \hline
 			  MP/s2    & $2\tiny{\times} 2$ \\ \hline
 			  C/s1     &$3\tiny{\times} 3 \tiny{\times} 32 \tiny{\times} 64$ \\ \hline
 			 MP/s2    & $2\tiny{\times} 2$ \\ \hline
 			  C/s1     &$3\tiny{\times} 3 \tiny{\times} 64 \tiny{\times} 64$\\ \hline
 			    
 			   MP/s2    & $2\tiny{\times} 2$ \\ \hline
 			   C/s1 & $3\tiny{\times} 3 \tiny{\times} 64 \tiny{\times} 128$\\ \hline 
 			  MP/s2    & $2\tiny{\times} 2$ \\ \hline
 			   C/s1 & $3\tiny{\times} 3 \tiny{\times} 128 \tiny{\times} 256$\\ \hline 
 			   Dropout & 0.25 \\ \hline
 		  FC/s1 & $256 \tiny{\times}$ nClasses \\ \hline
 			 SoftMax/s1 & Classifier \\ \hline
 			\end{tabular}
 	\end{center}
 	\caption{VGG-6 \protect\cite{sim14} variant used for case study 1. C is a convolutional layer and FC a dense layer; A BatchNorm and ReLU layer followed each C layer; MP denotes MaxPooling }  \label{tab:arch2} 
 \end{table}

We assess 3 distinct groups, each consisting of 3 layers, i.e., lower (around 2nd lowest conv layer), upper (around 2nd highest conv layer), and top layers (including last layer) for a malicious and non-malicious VGG-style CNN network.  We labeled about 2000 objects from the drone footage identified via a Faster-RCNN restricted to heights of [80,128] pixels to train the networks. 60\% of samples of each class became training data for the custom CNN. 40\% (but without labels) constitute the unlabeled dataset used by the investigator. Labels of this 40\% were used to compute evaluation metrics. The dataset with the number of samples per class is shown in Figure~\ref{fig:aicase}.
We considered two scenarios each with different datasets, i.e. $\mathcal{T}, \mathcal{U}$ and $\mathcal{U}',\mathcal{T}'$. Both $\mathcal{T}$ and $\mathcal{T}'$ (as well as $\mathcal{U}$ and $\mathcal{U}'$) had the same images but differed in their labeling. The dataset \emph{RefTrack} for non-malicious networks consists of two classes (referee and any other object). The \emph{RefTrack} corresponds to our described scenario. 
For the second dataset, \emph{AllTrack} attack objects and other detection objects are very similar. It consists of nine classes (all classes except the jersey number of the subject to attack). A malicious dataset has one more output class than a non-malicious dataset. The extra class is one of the five jersey numbers, e.g., 10, 11, 90, 4, 20. The \emph{AllTrack} dataset is designed to be more difficult for forensic investigation.

For non-malicious networks, the images of attack classes were assigned to either $\# home Away$ or $\# other Away$. We trained with image augmentation and class-weighing, yielding test accuracies consistently above 90\%. The confusion matrix showed that all classes were learned well. For each hyperparameter setting, we trained three models. 

For quantitative assessment, we introduce a measure denoted feature consistency $F_c$. It allows capturing to what extent a feature is associated with a suspicious concept (or class) only. We expect that malicious networks exhibit more features with high feature consistency for suspicious concepts than non-malicious networks. Feature consistency $F_c$ is computed on representatives $m_{j} \in G_{j}$ as the ratio of the number of samples $S \subseteq T_{j}$ that activate a particular feature or concept out of all samples $T_{j}$. It is expressed as a percentage, i.e., $F_c(m_{j}):= \frac{100|S|}{|T_{j}|}$. A suspicious concept is any of the jersey numbers 10, 11, 4, 90, and 20 (see classes in Figure~\ref{fig:aicase}). Figure~\ref{fig:feasamp} illustrates most activating test samples for three features from $G_{j}$ and their feature consistencies $F_c$. Thus, if for multiple groups $G_j$ feature consistency is large, the network is likely sensitive to the suspicious concept and, thereby, possibly malicious.

\begin{figure}[htp]	
		\centering	\centerline{
			\includegraphics[width=0.99\linewidth]{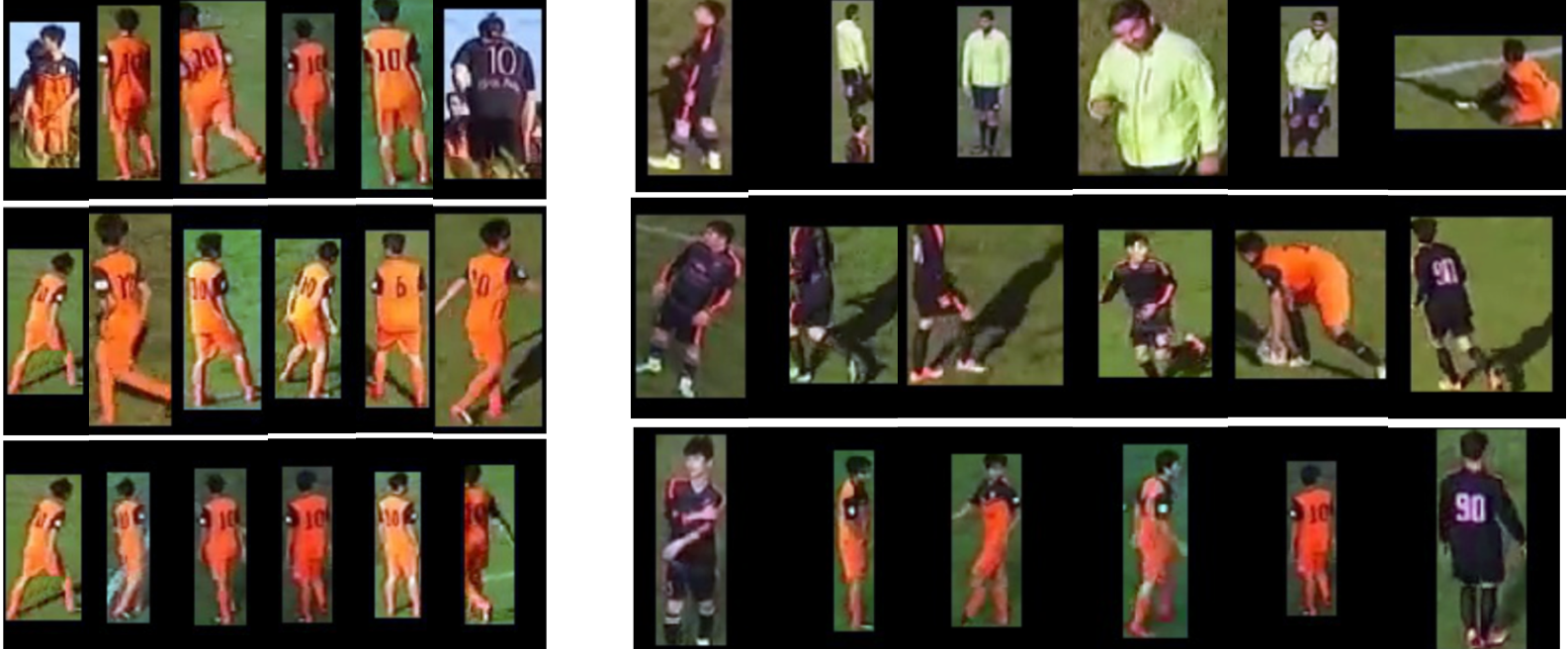}}
		\caption{Each row in the left and right panel shows top activating samples for a cluster representative $m_{l}$. Left: For the malicious network concept `10' is common among top activating samples. Right: For the non-malicious net, concept `10' is uncommon.} \label{fig:topds12} 
\end{figure}

\noindent \textbf{Results: } 
A qualitative assessment can be done using Figure \ref{fig:topds12} for the dataset \emph{RefTrack}. The figure shows the most activating, i.e., top $n=6$ samples, for a subset of all 500 cluster representatives for a malicious and non-malicious network trained with jersey number ``10 home'' as attack samples. Each row contains 6 samples $T_{j}$ for a specific $m_j$. An investigator checks each set of samples $T_j$ whether they consistently show the same suspicious characteristic, e.g., a jersey number. Some sets $T_j$ in Figure \ref{fig:topds12} for the malicious network show indeed consistent features that resemble the attack class, i.e., images with jersey number 10. In Figure \ref{fig:topds12} for the non-malicious network, no such sets can be identified. 

For quantitative assessment, Table~\ref{tab:feaConsAl} shows the average feature consistencies across all attack classes. An investigator aims to distinguish malicious from non-malicious networks. Malicious networks show higher feature consistencies for attack classes than non-malicious networks. Thus, they can be separated. Suppose an investigator looks at the top activating samples for the groups $G_j$ identified by our algorithm and illustrated in Figures \ref{fig:feasamp} and \ref{fig:topds12}. In that case, one will find groups where all samples share a malicious concept more often for malicious networks. The effectiveness depends on the chosen layer and the similarity of attack objects and non-attack objects. For a non-malicious network, feature consistency is 0 for all layers, indicating no features exist corresponding to attack objects, i.e., suspicious jersey numbers.  This holds although the training data consisted of objects of the incident (showing jersey numbers). But the suspicious concepts, i.e., jersey numbers, were not needed or beneficial for the task at hand, i.e., distinguishing between the referee and players. Thus, an investigator will not detect a suspicious concept in a group stemming from the non-malicious network. 
In contrast, malicious network results depend on the layer. There are no groups with a high feature consistency for lower layers. For other layers, a few percent of all features have a feature consistency of 1, indicating that the network learns features related to the incident. The top layers encode semantically rich information, and the lower layers are more basic, generic information common in many classes. Thus, the outcome is aligned with existing knowledge on DL. An investigator looking into all groups will detect such a group. One can be confident that the network was specifically trained to detect a suspicious concept.

 \subsection{Results for the second dataset for Case 1}
Feature consistencies are stated in Table \ref{tab:feaConsAl}. In the paper we discussed results for \emph{RefTrack}, showing that malicious networks are easy to identify.
The situation is more intricate for the dataset \emph{AllTrack} when attack classes and non-attack classes bear more similarities. More precisely, similar concepts are used to distinguish between classes of non-attack objects and attack and non-attack objects. In our case, jersey numbers are needed to distinguish among non-attack classes, and a jersey number (though a different one) is needed to identify the attack class. Such an overlap raises challenges, i.e., we observe that also the non-malicious networks contain some features with large feature consistencies. The malicious network still contains significantly more (A t-test gave a p-value $<$0.001). Malicious and non-malicious networks must discriminate between very similar classes, i.e., both have multiple classes focusing on jersey numbers. Thus, the investigation becomes more difficult if the attack objects are very similar to the classes that the network should detect. It is not sufficient to detect a group where all top samples share a suspicious concept. An investigator must compare the outcome of a potentially ``malicious'' system to an adequate reference, e.g., to a non-malicious system.

		\begin{table}[t]
			\centering \footnotesize
			\setlength\tabcolsep{2.5pt}
			 \begin{tabular}{c rr rr rr  rr  rr}
 		
& \multicolumn{2}{c}{Top Layers}&\multicolumn{2}{c}{Upper Layers}&\multicolumn{2}{c}{Lower Layers}\\ 
& Non-Ma.&Malic.&Non-Ma.&Malic.&Non-Ma.&Malic.\\ 

\multicolumn{1}{l}{RefTrack} &0.0\tiny{$\pm$ 0.0}&5.0\tiny{$\pm$ 1.8}&0.0\tiny{$\pm$ 0.0}&3.0\tiny{$\pm$ 0.7}&0.0\tiny{$\pm$ 0.0}&0.0\tiny{$\pm$ 0.0}\\ 
\multicolumn{1}{l}{AllTrack}&0.4\tiny{$\pm$ 0.4}&1.3\tiny{$\pm$ 0.6}&0.6\tiny{$\pm$ 0.5}&1.3\tiny{$\pm$ 0.5}&0.0\tiny{$\pm$ 0.0}&0.0\tiny{$\pm$ 0.1}
 			 \end{tabular}
 			 \caption{Feature consistency $F_c$ for non- and malicious nets} \label{tab:feaConsAl} 
		\end{table}

Thus, our method allows distinguishing malicious from non-malicious networks. 
Investigators can use the proposed method to identify groups from unlabeled data. They can compare the top samples and identify shared relevant concepts for the incident, i.e., if top samples share a suspicious concept, the network is likely malicious (see Figure \ref{fig:topds12}). While the method is widely applicable, as a limitation, it requires data related to the incident and manual work.



\subsection{Discussion Case 1}
On the technical side, the evaluation showed that malicious networks could be identified using visual investigation if different features are required to distinguish between non-attack objects and attack objects from non-attack objects. For the dataset \emph{RefTrack} the non-malicious network did not learn any features related to jersey numbers. However, they were shown in the training data, and they would, in principle, have proven helpful to distinguish the referee from the players. In contrast, for the dataset \emph{AllTrack} the non-malicious network learned such features since it had to distinguish between different jersey numbers. In the latter case, simply identifying a shared suspicious concept is not sufficient. Once a concept is identified, it must be shown that it is more frequent than in a non-malicious reference. This increases the investigator's effort since a reference is not readily available. An investigator might follow our evaluative process, including data labeling, training a malicious and non-malicious classifier (serving as reference), and comparing feature consistencies between networks as in Table \ref{tab:feaConsAl}. Furthermore, it is not necessarily evident what constitutes a suspicious concept. The op samples of a group might share multiple similarities. It depends on the investigator to identify those relevant to the incident. Thus, the investigator's judgment remains highly relevant. 

\subsection{Evaluation Case 2} 
We trained a VGG \cite{sim14} variant (Table \ref{tab:arch}) for 30 epochs without data augmentation with L2 weight regularization factor of 0.0005 with a batch size of 128 leading to test accuracies of about 45\%. Note that better-performing models would be to our advantage, i.e., to that of an investigator. Low accuracy might be used as a means to disguise. The reason is that low accuracy relates to noisy classifications making differences used for investigation between models (and features) less profound. For k-Means we used Python's sklearn library. We used stochastic gradient with learning rate decay starting from 0.12, decaying by a factor of 10 after epochs 20 and 30.\\

 \begin{table}[ht] 	
 	\begin{center}
 		\scriptsize
 		\setlength\tabcolsep{2.5pt}
	\centering
 		\begin{tabular}{|l| l |   }\hline
			Type/Stride& Filter Shape \\  \hline
			  C/s1     & $3\tiny{\times} 3 \tiny{\times} 3 \tiny{\times} 32$ \\ \hline
 			  MP/s2    & $2\tiny{\times} 2$ \\ \hline
 			  C/s1     &$3\tiny{\times} 3 \tiny{\times} 32 \tiny{\times} 64$ \\ \hline
 			 C/s1 & $3\tiny{\times} 3 \tiny{\times} 64 \tiny{\times} 64$ \\ \hline
 			 MP/s2    & $2\tiny{\times} 2$ \\ \hline
 			  C/s1     &$3\tiny{\times} 3 \tiny{\times} 64 \tiny{\times} 128$\\ \hline
 			   C/s1     & $3\tiny{\times} 3 \tiny{\times} 128 \tiny{\times} 128$\\\hline
 			   MP/s2    & $2\tiny{\times} 2$ \\ \hline
 			   C/s1 & $3\tiny{\times} 3 \tiny{\times} 128 \tiny{\times} 256$\\ \hline 
 			  C/s1     & $3\tiny{\times} 3 \tiny{\times} 256 \tiny{\times} 256$ \\ \hline
 			  MP/s2    & $2\tiny{\times} 2$ \\ \hline
 			   C/s1 & $3\tiny{\times} 3 \tiny{\times} 256 \tiny{\times} 512$\\ \hline 
 			  C/s1     & $1\tiny{\times} 1 \tiny{\times} 512 \tiny{\times} 512$ \\ \hline	
 		  FC/s1 & $512 \tiny{\times}$ nClasses \\ \hline
 			 SoftMax/s1 & Classifier \\ \hline
 			\end{tabular}
 	\end{center}
 	\caption{VGG-10 \protect\cite{sim14} variant used for case study 2. C is a convolutional layer and FC a dense layer; A BatchNorm and ReLU layer followed each C layer; MP denotes MaxPooling }  \label{tab:arch} 
 \end{table}

We trained on the Cifar-100 dataset. For $NT$, $RT$, and $ET$, we choose distinct sets of samples $S_O$ and $S_A$, each corresponding to those of a random class. Cifar-100 consists of 100 classes summarized by 20 more general categories, e.g., `people' contains baby, boy, girl, man, and woman. For the three classes $\mathcal{U}:=\{\zeta_O,\zeta_A,\zeta_R\}$ we use for $\zeta_A$ samples of a class from the same category as $S_A$, for $\zeta_O$ samples from the same category as $S_O$ and for $\zeta_R$ samples of a class not in any category of $S_O$ or $S_A$. We train the networks without the three classes $\mathcal{U}$. They are used for evaluation. We train 30 networks for each $NT$, $RT$, and $ET$. For each network, we randomly choose the sets $\mathcal{U}$.

\noindent \textbf{Results: Abnormal AI? RT?}  
The overall error $err(\mathcal{L})$ in Table~\ref{tab:gaet} is similar for $NT$, $RT$, and $ET$, indicating that manipulations do not distort overall performance. This is expected since only 1\% of training data differs for $NT$, $RT$, and $ET$. For individual classes, as expected, for $NT$ and $ET$ about 40\% of samples from $\mathcal{L}_A$ are classified as $A$, which means its error $err(\mathcal{L}_A)$ is roughly the error of the model $err(\mathcal{L})$ across all classes. Thus, both appear to be non-tampered. For $\zeta_A$ the error is large, indicating that for a non-tampered drone action $A$ is not triggered by the attacker's samples $S_A$. On contrary, for $RT$ samples $\zeta_A$ cause $A$, but $\mathcal{L}_A$ do not. 
Consequently, the tampering is easy to identify due to large error (similar to guessing) of $\mathcal{L}_A$.  Still, it is unknown what objects were used for tampering, i.e., set $S_A$.

	\begin{table}[t]
			\centering \footnotesize
			 \begin{tabular}{R{1cm}R{1.9cm}R{1.9cm}R{1.9cm}}
        Method	& $err(\mathcal{L})$ & $err(\mathcal{L}_A)$  & $C(\zeta_A,A)$  \\ 
$NT$ &0.443$\pm\tiny{0.01}$&0.401$\pm\tiny{0.18}$&0.991$\pm\tiny{0.01}$ \\
$RT$ &0.438$\pm\tiny{0.01}$&0.993$\pm\tiny{0.01}$&0.403$\pm\tiny{0.18}$ \\ 
$ET$ &0.44$\pm\tiny{0.01}$&0.421$\pm\tiny{0.17}$&0.432$\pm\tiny{0.16}$ \\ 
 			\end{tabular}
 			\caption{Accuracy for different tampering attempts.} \label{tab:gaet}  
 		\end{table}


\noindent \textbf{Results: NT or ET?} 
\noindent \emph{Identify $\zeta_R$ using misclassification analysis: } 
Table~\ref{tab:ransa} shows the output for assessing unlabeled samples $\mathcal{U}$. As expected, samples from $\zeta_O$ and $\zeta_A$ have a lower rank and lower error than random. However, the rank of the random class seems fairly low, i.e., under the assumption that on average a random class should have the mean rank of all classes, which is 48.5 (we trained using 97 classes). 
The reason is that class $A$ is trained with $\mathcal{T}_A:=S_A \cup S_O$, covering different concepts and being larger than any set $\mathcal{T}_i$ for any other class $i$. Thus, there are more concepts associated with class $A$. This makes it more likely that random samples $\zeta_R$ are classified as $A$. Standard deviations seem fairly large. This follows from the Cifar-100 dataset. Sub-classes in some categories exhibit limited similarity, while other categories overlap, e.g., `streetcar' and `pick-up truck' are in different categories, whereas `clock' and `keyboard'  are in the same category. Still, overall differences are significant (a t-test gave p-value $<0.0001$).

			

 	
\noindent \emph{Separate $\zeta_O$ and $\zeta_A$ using feature-based analysis:} Table~\ref{tab:ransa} shows the intersection of features of the two unlabeled data sets $\zeta_A$ and $\zeta_O$ with the features $F_{\mathcal{L}_A}$ obtained for $\mathcal{L}_A$ that is supposed to trigger $A$. It can be seen that while samples from $\zeta_A$ share many features with $\mathcal{L}_A$, set $\zeta_O$ does not. Thus, we can distinguish $\zeta_O$  from $\zeta_A$, leaving us with the attacker samples $\zeta_A$ that we aimed to identify. 

		\begin{table}[th]
			\centering \scriptsize
			 \setlength\tabcolsep{5pt}
			 \begin{tabular}{R{1.2cm}R{1.2cm}R{1.2cm}R{1.5cm}R{1.5cm}}
  $Rk(\zeta_O,A)$ & $Rk(\zeta_R,A)$ & $Rk(\zeta_A,A)$ & $|F_{\zeta_A} \cap F_{\mathcal{L}_A}|$ &      $|F_{\zeta_O} \cap F_{\mathcal{L}_A}|$ \\ 
2.0$\pm\tiny{6.61}$&12.0$\pm\tiny{17.15}$&4.0$\pm\tiny{9.58}$ &2.14&0.10
 			\end{tabular}
 			\caption{Median Rank for action $A$ for $\zeta \in \mathcal{U}$; Mean \# features in intersection with $\mathcal{L}_A$  } \label{tab:ransa} 
 	\end{table}

\subsection{Discussion Case 2}
Our evaluation highlighted that an attacker could easily associate an output with different stimuli, e.g., in our scenario, an attacker can trigger dropping a parcel on a person rather than on a heliport. However, this can also be identified.  In practice, the evaluation is more intricate since there are likely more than three sets of samples. However, they all fall into one of the three considered types. Thus, this should not hinder the application of our method. An adversary has multiple means for counter forensics, possibly at the price of making the attacks ``less targeted''. For example, the tampered system might be trained only briefly or with very few samples of the attack class. This could cause the system to only occasionally trigger the malicious action, although the attacker would want it to always trigger in this situation. But, in turn, detection is much more difficult, i.e., notable differences between the attacker dataset and other sets become more subtle.

\section{Case 3: Generalized Setting}

\begin{figure}[bth]	
		\centering	\centerline{\includegraphics[width=0.95\linewidth]{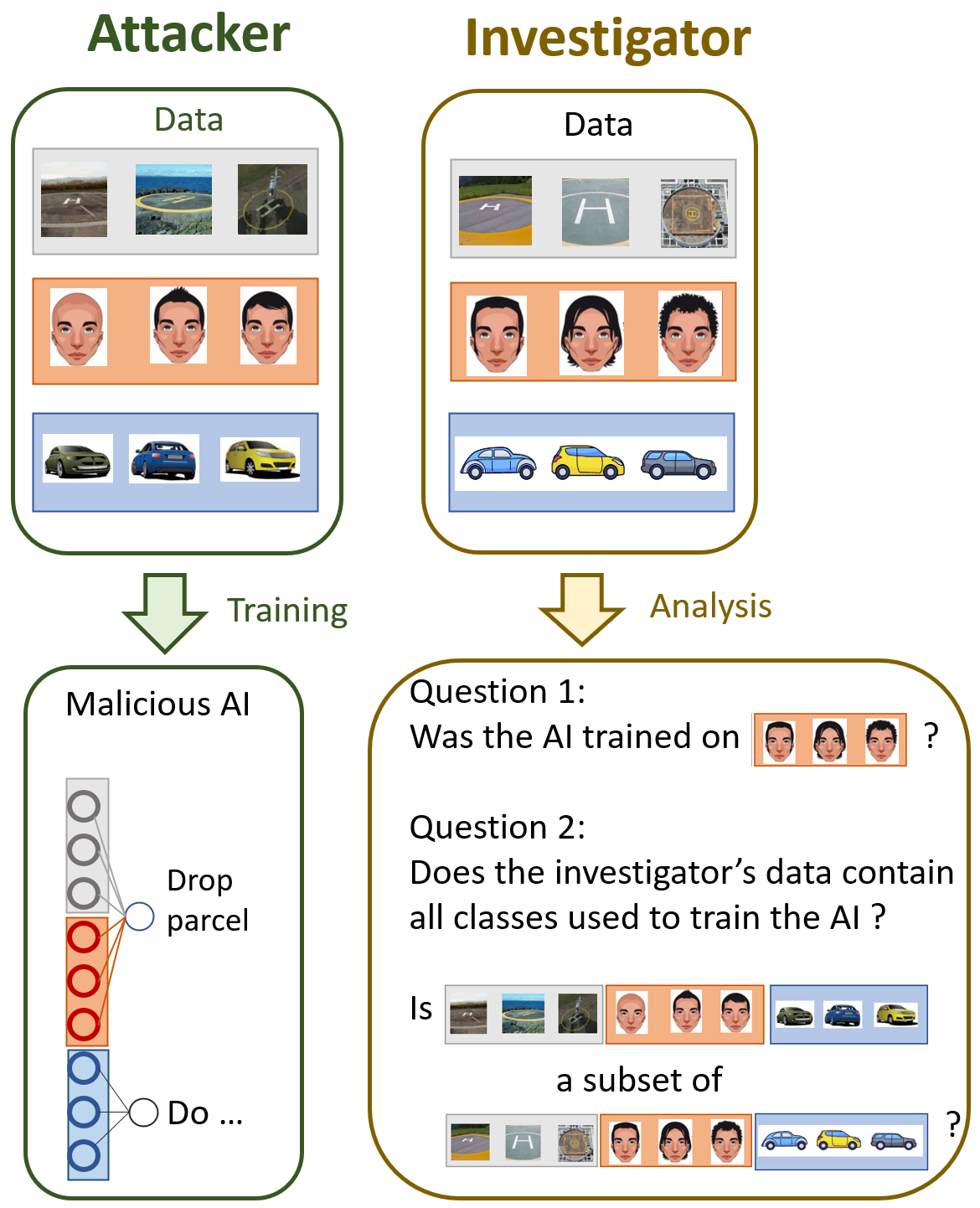}}
		\caption{General questions with respect to training data} \label{fig:case3}
\end{figure}

\noindent \textbf{Forensic Goals and Evidence: }  
The aim is to answer two general questions:\\
\noindent \emph{i) Was the network trained to detect a specific set of samples sharing similarities, e.g., constituting a class?}\\
\noindent \emph{ii) Does the investigative data contain all classes used for training a classifier?}\\
We assume that the investigator has access to data $\mathcal{D} \subseteq (\mathcal{U} \cap \mathcal{L})$. In particular, $\mathcal{D}$ contains some sets that are similar or even identical to some sets used for training the model $\mathcal{T}$. The two questions are illustrated in Figure \ref{fig:case3}.

For Question (i) the investigator wants to assess if a network was trained to identify samples $D\in \mathcal{D}$ as a class, i.e., if the network was  trained specifically to treat samples $D$ as belonging to one class $Y$. For example, a drone might be trained to identify a heliport for dropping cargo and a general class constituting obstacles. Still, it might not be trained specifically to identify humans, let alone specific individuals. Knowing whether the network recognizes a specific person can help judge a network as suspicious. In the prior section, we employed a semi-automatic approach that identified samples leading to the strongest activations for each neuron. Here, we present another methodology that relies on assessing mean activations of samples of a class, specifically, whether a network contains features that activate primarily for that class. 

Question (ii) is motivated by the observation that an investigator certainly does not want to miss relevant classes in her investigative data. If so, it is not fully known to what the network reacts and a network cannot be reliably judged as being (non-)malicious. 

\noindent\textbf{Investigation:}
A network learns features that allow discrimination among classes. While lower layers encode characteristics that are often common for many classes, upper layers contain more specific features that are  typically characteristic for one (or at most a few) classes. 
A feature $j \in [0,|M|-1]$ is characteristic of a set $D$ if activations $M_j(X)$ for $X \in D$ are frequently considerably larger than those of other sets $D'$. 
To capture this intuition we compute averages and compare them. We define the average activation for samples $D$ and feature $j$ as: \eq{\overline{M_j}(D):=\frac{\sum_{X \in D} M_j(X)}{|D|}}
The standard deviation s \eq{sd(M_j(D)):=\frac{\sqrt{\sum_{X \in D} (M_j(X)-\overline{M_j}(D))^2}}{|D|}}

To address Question (i), let $D_{all}:= \cup_{S \in \mathcal{D}} S$ be the union of all samples used for investigation.
We say a feature $j$ is characteristic for samples $D$, if \eq{\overline{M_j}(D)>\overline{M_j}(D_{all})+c_1\cdot sd(M_j(D_{all}))}, where $c_1$ is a fixed constant, i.e., we shall use 1.5. We define the number of characteristic features $N(D)$ as: 
\eq{N(D):=&|\{i|i \in [0,|M|-1], \overline{M_j}(D)>\overline{M_j}(D_{all})\\
&+c_1\cdot sd(M_j(D_{all}))\}|}
A set of samples $D$ was not used for training if there are fewer characteristic features $N(D)$ than for sets that are known to be used for training (or are very similar to such sets) $N(D')$.
The average number of activated features for sets $\mathcal{D}$ is: 
\eq{\overline{N}(\mathcal{D}):=\frac{\sum_{S \in \mathcal{D}} N(S)}{|\mathcal{D}|}}
The standard deviation is \eq{sd(N(\mathcal{D})):=\frac{\sqrt{\sum_{S \in \mathcal{D}} (N(S)-\overline{N}(\mathcal{D}))^2}}{|\mathcal{D}|}}
We say that samples $D$ have not been used for training if $N(D) + c_2\cdot  sd(N(\mathcal{D})) <\overline{N}(\mathcal{D})$, where $c_2$ is a constant, i.e., we used $c_2=1.5$. We discuss finding the thresholds in our evaluation.

To address Question (ii), the idea is to assess if the network contains features that never strongly activate for any of the sets of the investigative data. This question is more challenging than Question (i) since average activations of two features can vary strongly, even when using the actual training data $\mathcal{T}$ since some features (or patterns) tend to be more common among samples in general than others. Thus, detection tends to be more brittle. 
We aim to identify features that are not activated for any of the sets in the investigative data.
We compute the maximum mean activation of a feature across all sets $D \in \mathcal{D}$, i.e., $M_{j,max}(\mathcal{D})=\max_{D \in \mathcal{D}} \overline{M_j}(D)$.
We consider a feature as non-activated (for all sets) if it is below a threshold $c_3$. We assessed two values of $c_3$, i.e., 0.025 and 0.05. We discuss finding the thresholds in the evaluation. The number of non-activated features for the investigative data is as follows:
$NA(\mathcal{D}):=\{j|M_{j,max}(\mathcal{D})<c_3 \}$

We say that the investigative data $\mathcal{D}$ misses at least a set $D\in \mathcal{T}$, if $NA(\mathcal{D})>c_4$, where $c_4=3$ is a threshold.

\subsection{Evaluation Case 3} 

\noindent\textbf{Setup.} We used two classifiers, i.e., VGG-11 (V11), and ResNet-10 (R10). We employed two datasets, namely Fashion-MNIST and Cifar-10. Fashion-MNIST consists of 70k 28x28 images of clothing stemming from 10 classes. As data preprocessing for Fashion-MNIST, we scaled all images to 32x32 and performed standardization. We assume that activations $M(X)$ stem from the second last layer before the linear layer, i.e., consisting of 512 neurons. We also investigated using three layers, i.e., the fifth last layer up to the second last layer, but we do not include detailed results but discuss their differences qualitatively. 
To investigate Question (i) we train a classifier on $\mathcal{T'}=\mathcal{T}\setminus D$, i.e., with all but one class $D \in \mathcal{T}$. We assume investigative data $\mathcal{D}=\mathcal{T}$. Thus, our method correctly detects the non-trained class corresponding to samples $D$, if (a) it says the classifier was not trained on samples $D$ and it says that the classifier was trained on all other classes $\mathcal{T'}$. 

In our evaluation, we leave out each class $D \in \mathcal{T}$ and train per left-out class three classifiers of each classifier type, i.e., V11 and R10. That is, we trained a total of 60 classifiers for the 10 classes. We report the mean detection accuracy and the standard deviation as well as the model accuracy and standard deviation though this is not a primary concern.
To investigate Question (ii) we could train using all classes $\mathcal{T}$. However, to get more diversity in outcomes and show that our method works robustly, we reuse more diverse training data, i.e., we reuse the same classifiers as for Question (i), i.e., trained on $\mathcal{T'}=\mathcal{T}\setminus D$. The investigative data lacks one additional class $D'$, i.e., $\mathcal{D}=\mathcal{T'}\setminus D'$ with $D' \in \mathcal{T'}$. We let each class $D' \in \mathcal{T'}$ be missing once and say that the non-completeness of $\mathcal{D}$ was correctly detected, if $NA(\mathcal{D})>c_4$, i.e., the algorithm outputs that the investigative data lacks a class $\mathcal{D}$ and it outputs that actual training data $\mathcal{T'}$ does not lack a class. 

\smallskip

\begin{figure*}[ht]	
		\centering	\centerline{
			\includegraphics[width=0.99\linewidth]{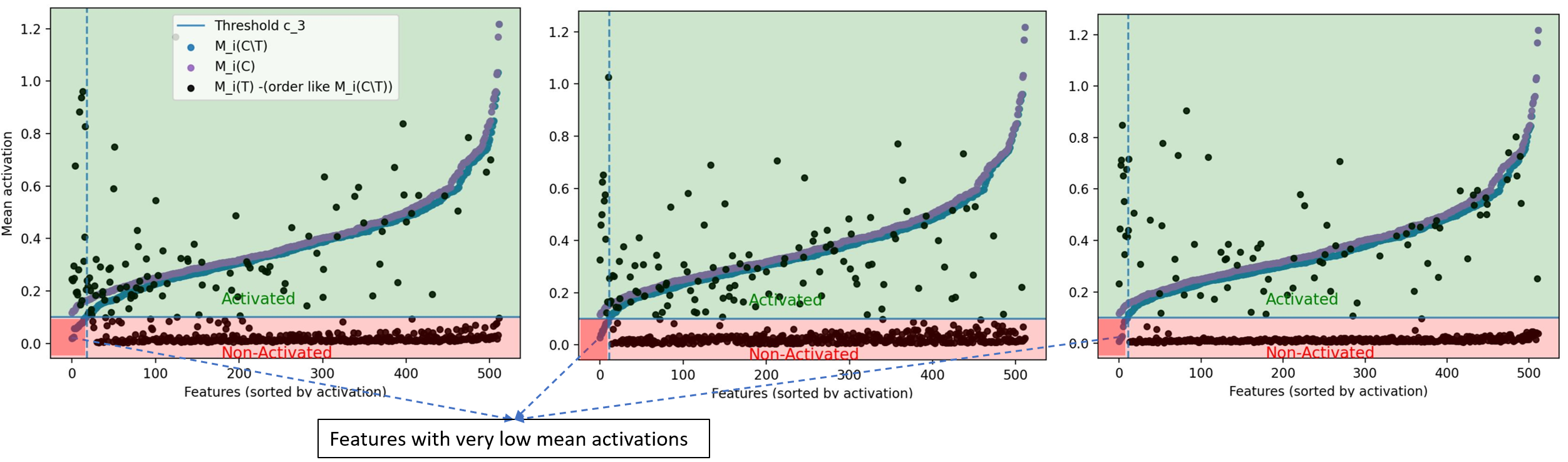}}
		\caption{Mean activations sorted for all data and data missing a class for V-11 and Cifar-10. In panel $j$ class $j$ with samples $T$ is not part of the investigative data ($D=C\setminus T$). For $D$ there tends to be more features that have very low mean activations (compared to the full data $C$). For the class $T$ only few features show large mean activations.} \label{fig:miss}
\end{figure*}

\noindent \textbf{Results Question (i):} The threshold constants $c_1$  and $c_2$ can be identified through exploratory analysis, i.e., doing plots like Figure \ref{fig:miss} shown in our results for different values for $c_1$ and coloring activated and non-activated features, i.e., increasing $c_1$ so that only features with large means are characteristic for a class.\footnote{One might also apply a threshold directly on the mean for a feature by looking at the plot.} Note, we do not use any information on the set $D$ we aim to assess for that purpose. Qualitatively, in Figure \ref{fig:miss} it becomes apparent that if a class is left out from training, few or no mean activations of features tend to be large. In Table \ref{tab:notr} we can see that we can correctly detect the non-trained class (among all classes) in the majority of cases -- even using the same parameters $c_1$ and $c_2$. 
This procedure works well also in the scenario, where activations $M$ consist of three (different) layers. If activations of multiple layer types are merged, it generally suffices to focus on neurons with large mean activations, which tend to be those after batchnorm, i.e., means of conv-layer tend to be near 0. If neurons of different layers are approximately of the same magnitude, they do not have to be distinguished and can be analyzed jointly.

\begin{figure*}[ht]	
		\centering	\centerline{
			\includegraphics[width=0.99\linewidth]{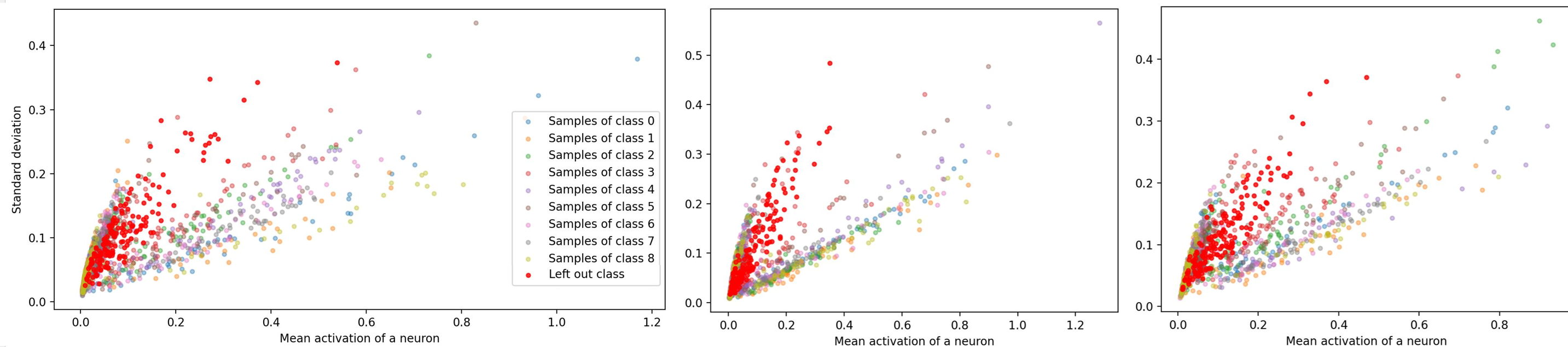}}
		\caption{Mean activations and their standard deviations of features for different classes for V11 and Cifar-10. In panel $i$, only class $i$ (red) was left out, i.e., not used for training. Classes not used for training have few or no features with large mean activation.} \label{fig:notr}
\end{figure*}

		\begin{table}[th]
			\centering \scriptsize
			 \setlength\tabcolsep{5pt}
			 \begin{tabular}{|l|l|l|l|l|} \hline
			 Network & Dataset &  Model Acc.  & Detect Acc. \\  \hline

 R10&Cifar-10&0.88\tiny{\text{$\pm$}0.01}&1.0\tiny{\text{$\pm$}0.0}\\ \hline
 R10&FashionMNIST&0.94\tiny{\text{$\pm$}0.011}&1.0\tiny{\text{$\pm$}0.0}\\ \hline
 V11&Cifar-10&0.81\tiny{\text{$\pm$}0.014}&0.9\tiny{\text{$\pm$}0.3}\\ \hline
 V11&FashionMNIST&0.93\tiny{\text{$\pm$}0.013}&0.7\tiny{\text{$\pm$}0.458}\\ \hline

 			\end{tabular}
 			\caption{Model Accuracy and detection accuracy for a non-trained class} \label{tab:notr} 
 	\end{table} 
\noindent \textbf{Results Question (ii):}
To find thresholds $c_3$ and $c_4$, we performed a plot like in Figure \ref{fig:miss}, but intentionally leaving out a set $D' \setminus \mathcal{D}$. If the number of features with very low activations increased significantly, i.e., among the lowest activating features, the majority intersects with activated features of set $D'$ (see Figure \ref{fig:miss}) then it can be concluded that the investigative data is complete. If only a small fraction, i.e., less than 1/2, of the lowest activating features stem from the left $D'$ then many other features have not been activated for the investigative data. The challenge is that some features can have low mean activation for all classes. Thus, among the lowest activating features there can be some features that are activating for a set in $\mathcal{D}$, but the activation is small.

In our analysis, we aim to distinguish between the case that the dataset $\mathcal{D}$ is complete, i.e., it is equal to the training data, or at least one set $D' \in \mathcal{D}$ is missing. A set is missing if the number of non-activated features is above a threshold $c_4$ that captures the number of anticipated ``outliers'', i.e., features that have very low activation although they are part of the data $\mathcal{D}$. As can be seen in Figure \ref{fig:miss}, this approach works well if we have access to activations of a specific layer, but noise significantly increases if we assume that activations stem from multiple layers. The separation of activating and non-activating features based on means is a challenge. The good thing is that means of features of different layers of a conv-layer, ReLU, and a batchnorm layer tend to differ. The problem is that there is still overlap making a simple separation difficult. Under the assumption that activations are randomly permuted, this is non-trivial.

As can be seen in Figure \ref{fig:miss}, the mean activations fluctuate depending on the missing class. Still, for all left-out classes during training, there are multiple non-activated features below the threshold (blue dots) in the resulting networks. If we train on all data (violet dots), we can see that in particular, towards the lower end, there are clear differences between the blue dots. Furthermore, while some features that strongly activate for the missing class (black dots) also yield fairly strong activations for other classes, most features remain non-activated if the mean is large for some other class. Quantitatively, Table \ref{tab:Miss} shows that accuracy tends to be high across classifiers, but they are  sensitive to the threshold $c_3$.

		\begin{table}[th]
			\centering \scriptsize
			 \setlength\tabcolsep{5pt}
			 \begin{tabular}{|l|l|l|l|l|} \hline
			 Net. & Dataset & Thres. $c_3$ & Model Acc. &  Detect Acc \\ \hline
R10&Cifar-10&0.05&0.88\tiny{\text{$\pm$}0.011}&0.8\tiny{\text{$\pm$}0.4}\\ \hline
R10&Cifar-10&0.025&0.88\tiny{\text{$\pm$}0.011}&0.46\tiny{\text{$\pm$}0.217}\\ \hline
R10&FashionMNIST&0.025&0.94\tiny{\text{$\pm$}0.011}&0.87\tiny{\text{$\pm$}0.34}\\ \hline
R10&FashionMNIST&0.05&0.94\tiny{\text{$\pm$}0.011}&0.33\tiny{\text{$\pm$}0.471}\\ \hline
V11&Cifar-10&0.05&0.81\tiny{\text{$\pm$}0.014}&0.92\tiny{\text{$\pm$}0.121}\\ \hline
V11&Cifar-10&0.025&0.81\tiny{\text{$\pm$}0.014}&0.13\tiny{\text{$\pm$}0.11}\\ \hline
V11&FashionMNIST&0.05&0.93\tiny{\text{$\pm$}0.013}&1.0\tiny{\text{$\pm$}0.0}\\ \hline
V11&FashionMNIST&0.025&0.93\tiny{\text{$\pm$}0.013}&0.91\tiny{\text{$\pm$}0.078}\\ \hline
 			\end{tabular}
 			\caption{Model accuracy and detection accuracy for a (sample) set contained in training but missing from investigative data  } \label{tab:Miss} 
 	\end{table} 

\subsection{General Discussion and Future Work}
Our conceptualization provides a rich set of options to investigate. Still, this set of options is by no means comprehensive, and it cannot be so since forensics and attackers are both constantly evolving. Additional questions for research include ``How can suspects based on other AI techniques such as reinforcement learning, other scenarios (see Figure \ref{fig:aifor}) and other deep learning network architectures such as LSTMs be identified?'', ``How can explainability methods be leveraged (modified) to support analysis of grey-box models ?'', ``How can data mining techniques, e.g., to identify anomalous decisions be leveraged ?'', ``How can other types of evidence such as operational data  be used?''. 

Our work is only a first step towards AI forensics. We demonstrated that perpetrators could often be identified in the given attack scenario. We showed how properties of AI systems, e.g., memory access patterns of deep learning and knowledge of alleged behavior (encoded via a public dataset), can be leveraged for forensics.
Our case studies are just a small selection of countless options. Furthermore, various design decisions must also be made for these options.  This naturally raises questions with respect to generalizability. We believe that our techniques are widely applicable since systems for very different applications use the same deep learning technology, e.g., object detection is used in self-driving cars, manufacturing, healthcare, etc. Most effort for investigation might not be related to applying our methods but to collecting adequate data, which holds true for many machine learning projects. For example, an investigator might rely on data similar to what the system observed during the incident to understand what caused the AI to act in a certain way. However, getting such data might be difficult and costly. Even if data is acquired, it might differ from the incident in subtle but important ways. That is, the acquired data might not trigger malicious actions during the incident due to differences in data. To this end, evaluating more DL architectures on more datasets might be beneficial, but this alone is not sufficient given the many decisions to be made for any case study. More foundations are needed in deep learning in general, and more studies are needed that are only dedicated to a specific forensic case or question.  \\

Our work is not without limitations. For once, both our image classification networks seem fairly inaccurate. Note that our proposed techniques for forensic investigation become more reliable given higher accuracy since this reduces noise. As of today, in safety-critical applications like self-driving cars, outputs of computer vision components are used in conjunction with other sensors. Thus, an attack on a real system might be more complex than just retraining a single classifier. Still, modifying a classifier like a computer vision component seems an important step. Furthermore, one might question whether an attacker would attempt to manipulate an AI rather than just perform physical modifications to a device or adjust control logic in other ways. To this end, there is no definitive answer yet. We believe there are multiple decisive factors: i) Maturity of AI forensics. The availability of methods to identify attacks is making attacks less attractive. On the contrary, if law enforcement is not prepared to investigate an AI, it becomes more attractive for attackers to manipulate AI; ii) The evolution of AI. AI is likely to become easier to use, more autonomous, and cover more applications. We believe that this will make it more attractive to manipulate AI. iii) Regulation of AI. If AI systems are regulated to follow specific design guidelines (that also ease forensic investigation), this might make AI more difficult to abuse. In summary, for some time to come, AI might not be designed to be malicious, but the risk for this to eventually happen is large, and, thus, forensic work should start now to be one step ahead of attackers.


The overarching approach of our methods was to identify data samples that trigger malicious actions and present data samples having similar activations to the investigator. Our methods emphasize that the role of the digital forensic investigator remains highly important. In particular, for our first case study, a forensic investigator must identify concepts related to the incident, label data, etc. While some of these tasks might be further automated, we believe that the black-box nature of AI combined with its novelty makes the role of forensic investigators more difficult and more relevant in the context of AI.\\

Rather than working on data samples, one might also work with features, e.g., reconstructing features of a neural network.  There is a rich set of explainability methods \cite{arr20,sch19} that might be leveraged as part of the investigative process. Many common explainability methods such as LIME or SHAP that compute proxy models assuming black-box access or methods that require gradients such as GRAD-CAM can be helpful. However, methods for black-box models might not give the best results for grey-box models, since they do not leverage internal information. Furthermore, outputs (used by the system) might not always be easy to access, e.g., due to disguise layers as shown in the architecture for one of our case studies. Methods for white-box models cannot be applied directly since they often require internal information, e.g., gradients that cannot be computed. Still, adjusting XAI methods to the grey-box model might be helpful. For example, \cite{sch22con} proposed to decode layer activations. Potentially, this might allow identifying concepts of the input a classifier reacts to without the need for gradients. Since this method was developed assuming access to a single layer, more work is needed to clarify whether this or other techniques work when exact access to a single layer is unavailable.\\

\section{Conclusions}
We conceptualized AI forensics and identified malicious AI systems as part of forensic work based on two case studies leveraging layer activations in the grey box model. Forensic work poses novel challenges to investigators, and AI systems provide ample opportunities for hiding evidence, such as obfuscating computational patterns.  Techniques from emerging areas in AI, such as reverse engineering, explainability, adversarial analysis, and testing,  play an integral part in the investigation, but often require adjustment for forensic work. Thus, the `cat-and-mouse' game between attackers and forensic experts has just begun. 

\bibliographystyle{acm}
\bibliography{refs}

\section{Appendix}

\subsection{Drone design options}
\textbf{On-board processing vs.~communicating to server}: A key challenge for drones, in general, is limited battery power. From a system design perspective, two fundamental design options strongly impact energy consumption: Transmit the image to a server or process it on the drone. While both options have their strengths and weaknesses, in the paper, we discussed the latter, more battery energy-consuming option, because arguably, a non-autonomous drone would leave digital traces if it had to communicate with the nearest cell tower or WLAN access point. This poses a high risk for an attacker since the server the drone communicates with might be identifiable. 
Furthermore, if the server does a similar type of processing as the autonomous drone would, the investigative methods are the same once it is identified. From an investigative point of view, where the emphasis is on analyzing the AI system, the design options are the same. Also, both are found in practice.\\
\textbf{Object tracking vs. object recognition}:
We employ object recognition, assuming that each image of a drone is analyzed independently. There exist techniques designed for object tracking. However, these networks also rely on CNN layers, and they are not necessarily advantageous, e.g., in terms of ease of use and computational needs.
\textbf{Additional Custom CNN vs. Fine-Tuning of object detector}: For the drone design, one might fine-tune an object detector or use a custom CNN to refine the output of the object detector. The former has slightly lower energy consumption for the drone, but arguably requires more effort. We opted for a custom CNN, but both options are very similar -- also from an investigative perspective.\\
\textbf{Choice of Object Detector}: We opted for Faster-RCNN, but acknowledge that any other object detector could be used and not a key factor for our analysis. For example, YOLO provides a computationally lightweight object detector. Also, our neural network designs do not require much computing (our custom CNN only has a few layers). However, our networks might be further optimized toward less energy consumption. Furthermore, other architectures like MobileNet might lead to even lower energy consumption, though potentially at the cost of reduced detection/classification performance.\\
\textbf{Drone control logic}: Controlling a drone is complex. Therefore, our control logic consisting of a few simple rules is on a rather high level. The reason is that drone control algorithms are not the focus of investigation and, thus, are not of primary interest, though one could envision other scenarios where they are.
\textbf{Tracking referee vs. ball}: From a conceptual point of view, there is little difference. Also, the drone image covers a fairly large area of the sports ground (much larger than on the cropped images in the paper). Therefore, it is rarely the case that any action is missed if the focus is on the referee.
\textbf{Disguise and Impact on investigator}: Another reason why an investigator might not rely only on the last outputs based on the order of computation is disguise layers as shown in one of the figures in the paper: the network might intertwine genuine and non-genuine layer computations, where the last layers might consist of non-genuine layers. Thus, the investigator has to assess all layers.


\subsection{Dataset(s) for Case 1}
\emph{Data labeling}:   The dataset was collected by extracting images from the drone video lasting about 35 minutes. For a frame of the video, we obtained all objects (as returned by faster RCNN). We retained those with heights between 80 to 128 and labeled them according to the 10 classes shown in the figure in the paper. Objects with lower heights are hard to assess visually. After processing each frame, we skipped 50 frames to have some variation between frames and reduce the overall labeling load. This left us with 2013 labeled samples. After a first labeling round, we checked for mis-labelings and fixed those we identified.


\emph{Data source}: The data stems from a public event in a public setting uploaded on YouTube, where spectators are welcome, and filming is commonplace. We are happy to share more information, including our labeled dataset, or upload it publicly upon reviewer request. We only show very low-resolution images of players in the paper to respect their privacy, though millions of YouTube videos covering more private themes have been used in research, e.g., see the 8 million video \emph{YouTube-8M} dataset. We also contacted the uploader and carefully took into account ethical pros- and cons\footnote{e.g., as in \url{https://slate.com/technology/2019/06/youtube-twitter-irb-human-subjects-research-social-media-mining.html}}.

\end{document}


\maketitle

\section{Appendix}
The Appendix contains extra material that is either illustrative (e.g., additional figures and summary tables) or provides details or clarifications that are not central to the paper but of interest to reviewers.

\section{AI vs Classical Software} \label{sec:aisoft} 
A summary of differences is provided in Table~\ref{tab:aisoft} and those not described in the manuscript are further detailed below.

\section{Malicious by Design} \label{sec:probAp}

\subsection{Strategies for Manipulation} \label{sec:Manu}
Attacks based on the \emph{model} itself might also be conducted. For example, parameters of a CNN can be changed so that a system fails to recognize specific objects.
For attacks based on \emph{AI's objectives}, an attacker might change a model's objective (used during training). For example, a system that is supposed to be fair due to a regularization term in the objective (like \cite{kam11}) might be trained without it since fairness might be at odds with other performance measures. 
An \emph{AI system might learn continuously} \cite{par19} where attackers can manipulate the learning mechanism. For example, she might cause intentional (catastrophic) forgetting \cite{par19} of the original task.



\section{AI Forensics}

\subsection{Second dataset for Case 1} \label{sec:secdat}
The second dataset \emph{AllTrack} consists of nine classes (all classes except the jersey number of the subject to attack). A malicious dataset has one more output class than the non-malicious dataset. The extra class is one of the five jersey numbers, e.g., 10, 11, 90, 4, 20. The \emph{AllTrack} dataset is designed to be more difficult for forensic investigation. In this dataset, attack objects and other detection objects are very similar.

 \subsection{Results for second dataset for Case 1}
Feature consistencies are stated in Table \ref{tab:feaConsAl}. In the paper we discussed results for \emph{RefTrack}, showing that malicious networks are easy to identify.
The situation is more intricate for the dataset \emph{AllTrack}, when attack classes and non-attack classes bear more similarities. More precisely, if similar concepts are used to distinguish between classes of non-attack objects and attack and non-attack objects. In our case, jersey numbers are needed to distinguish among non-attack classes, and a jersey number (though a different one) is needed to identify the attack class. Such an overlap raises challenges, i.e., we observe that also the non-malicious networks contain some features with large feature consistencies. The malicious network still contains significantly more (A t-test gave a p-value $<$0.001). Malicious and non-malicious networks must discriminate between very similar classes, i.e., both have multiple classes focusing on jersey numbers. Thus, the investigation becomes more difficult if the attack objects are very similar to the classes that the network should detect. It is not sufficient to detect a group where all top samples share a suspicious concept. An investigator must compare the outcome of a potentially ``malicious'' system to an adequate reference, e.g., to a non-malicious system.

		\begin{table}[t]
			\centering \footnotesize
			\setlength\tabcolsep{2.5pt}
			 \begin{tabular}{c rr rr rr  rr  rr}
 		
& \multicolumn{2}{c}{Top Layers}&\multicolumn{2}{c}{Upper Layers}&\multicolumn{2}{c}{Lower Layers}\\ 
& Non-Ma.&Malic.&Non-Ma.&Malic.&Non-Ma.&Malic.\\ 

\multicolumn{1}{l}{RefTrack} &0.0\tiny{$\pm$ 0.0}&5.0\tiny{$\pm$ 1.8}&0.0\tiny{$\pm$ 0.0}&3.0\tiny{$\pm$ 0.7}&0.0\tiny{$\pm$ 0.0}&0.0\tiny{$\pm$ 0.0}\\ 
\multicolumn{1}{l}{AllTrack}&0.4\tiny{$\pm$ 0.4}&1.3\tiny{$\pm$ 0.6}&0.6\tiny{$\pm$ 0.5}&1.3\tiny{$\pm$ 0.5}&0.0\tiny{$\pm$ 0.0}&0.0\tiny{$\pm$ 0.1}
 			 \end{tabular}
 			 \caption{Feature consistency $F_c$ for non- and malicious nets} \label{tab:feaConsAl} 
		\end{table}

 \section{Discussion}


\bibliographystyle{named}
\bibliography{refs}